\documentclass{article}

\usepackage[final, nonatbib]{neurips_2024}
\usepackage[numbers]{natbib}

\usepackage[utf8]{inputenc} 
\usepackage[T1]{fontenc}    
\usepackage{hyperref}       
\usepackage{url}            
\usepackage{booktabs}       
\usepackage{amsfonts}       
\usepackage{nicefrac}       
\usepackage{microtype}      
\usepackage{xcolor}         

\usepackage{graphicx}
\usepackage{multirow}
\usepackage{colortbl}
\usepackage{tabularx}
\usepackage{amsmath}

\newcommand\rot[1]{\rlap{\rotatebox{45}{#1}}}

\setcitestyle{square}

\title{DAug: Diffusion-based Channel Augmentation for \\Radiology Image Retrieval and Classification}

\author{%
Ying Jin$^{1,2}$ \quad Zhuoran Zhou$^1$ \quad Haoquan Fang$^1$ \quad Jenq-Neng Hwang$^1$ \\
$^1$University of Washington \quad $^2$Microsoft\\
\texttt{\{jinying, hqfang, hwang\}@uw.edu}, xcharxlie@gmail.com
}

\begin{document}

\maketitle

\begin{abstract}
  Medical image understanding requires meticulous examination of fine visual details, with particular regions requiring additional attention. While radiologists build such expertise over years of experience, it is challenging for AI models to learn where to look with limited amounts of training data. This limitation results in unsatisfying robustness in medical image understanding. To address this issue, we propose Diffusion-based Feature Augmentation (DAug), a portable method that improves a perception model's performance with a generative model's output. Specifically, we extend a radiology image to multiple channels, with the additional channels being the heatmaps of regions where diseases tend to develop. A diffusion-based image-to-image translation model was used to generate such heatmaps conditioned on selected disease classes. Our method is motivated by the fact that generative models learn the distribution of normal and abnormal images, and such knowledge is complementary to image understanding tasks. In addition, we propose the Image-Text-Class Hybrid Contrastive learning to utilize both text and class labels. With two novel approaches combined, our method surpasses baseline models without changing the model architecture, and achieves state-of-the-art performance on both medical image retrieval and classification tasks.
\end{abstract}

\section{Introduction}
\label{sec:intro}

Shortages and burnout of radiologists are significant problems worldwide and leave risks to patient care \cite{cao2023current, rimmer2017radiologist}. Training a radiologist takes thirteen to fifteen years \cite{years_radiologist}, making AI models assisting in diagnostics a scalable solution. Notably, Chest X-ray (CXR) classification and retrieval are fundamental problems that have solid clinical values, as classification can cross-check with doctors, and retrieval allows comparison with historical cases for more accurate diagnoses.

In recent years, the performance of various image understanding tasks are meaningfully improved by leveraging pretrained vision models. Contrastive models like CLIP \cite{radford2021learning} provide a strong baseline for understanding the common, scenic images. However, when transferred to medical images, their pretrained capabilities are under-explored for two reasons. First, medical images require meticulous examination of fine details, which differs from scenic images where salient objects are large and have clear boundaries. Second, radiology images like Chest X-rays are monochrome, preventing full utilization of the pretrained model's capability to utilize all three color channels. Consequently, vision models trained on limited radiology images may not attend to the correct regions for accurate understanding. Previous work \cite{yang2023set} has shown that drawing visual marks on the input image improves model performance on tasks related to visual grounding. Therefore, we are motivated to explore highlighting potential disease areas in medical images to improve the performance of medical imaging tasks. In this study, we propose a new method to achieve this goal by adding visual grounding hints as additional image channels. Specifically, we augment the image feature with abnormality heatmaps as additional channels alongside the original monochrome medical image. The augmented feature provides additional information to the model on areas requiring extra attention, mimicking the routine of experienced radiologists. This method, dubbed Diffusion-based Feature Augmentation (DAug), is portable onto a wide range of model architectures and leverages their native compatibility with multi-channel (RGB) images. An example of such heatmaps (e.g., for the disease of cardiomegaly) can be found in \autoref{fig:heatmap}.

\begin{figure*}[t]
  \centering
  \includegraphics[width=.9\linewidth]{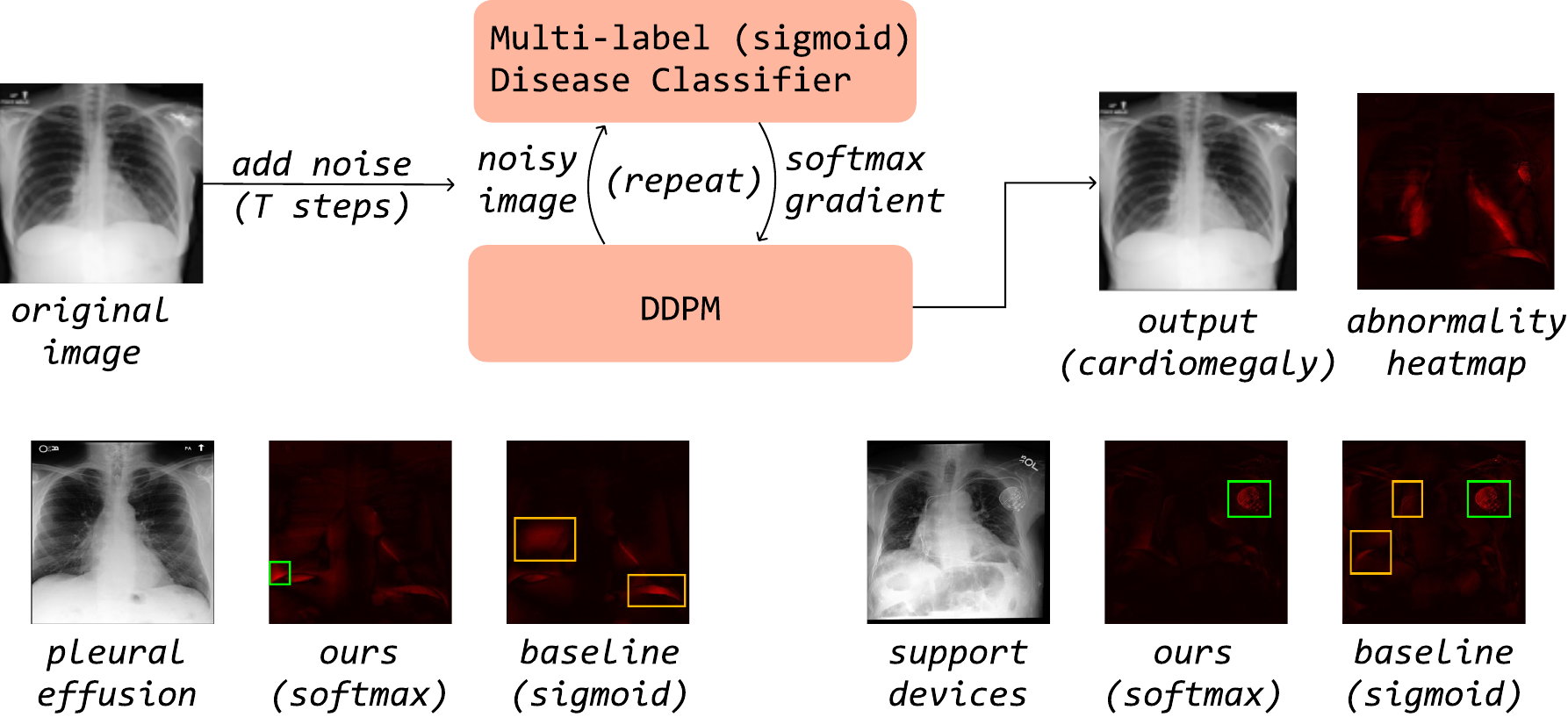}
  \caption{Diffusion-based Feature Augmentation (DAug) pipeline. The original image is translated into a diseased or healthy version with a classifier-guided diffusion model. The upper row shows an example where an image with a healthy heart is turned into a cardiomegaly (enlarged heart) version. The difference between the input and output images produces a heatmap highlighting the potential area of the corresponding disease. In the case of cardiomegaly, the heatmap correctly highlights the boundary of the heart. The heatmaps are added to the original monochrome radiology image as additional image channels, resulting in an augmented input feature that can improve the performance of downstream tasks. The DAug features support multiple disease categories (two examples in the second row), and our softmax-based approach generates more accurate heatmaps than the existing baseline. Green and orange bounding boxes indicate correctly and wrongly highlighted regions, respectively.}
  \label{fig:teaser}
\end{figure*}

Concretely, we train an image-to-image translation model to convert an input image to heatmaps of selected disease classes. Each heatmap highlights the region where the disease could potentially occur. The heatmap, as an additional feature channel, explicitly directs the model's attention to clinically significant areas, a skill difficult to acquire through conventional model training methods. The image-to-image translation model is implemented by repurposing a classifier-guided diffusion model \cite{dhariwal2021diffusion}. First, we add Gaussian noise into a Chest X-ray image, making a noised version of the original image as the model input. Second, the diffusion model, guided by a disease classifier, removes the noise toward a direction where a disease is mitigated or exacerbated. In \autoref{fig:heatmap}, we show an example where cardiomegaly is worsened in the diffusion output. The difference between the original image and the output yields an attention heatmap that highlights the potential disease area (e.g., the boundaries of the heart for cardiomegaly). Through this generative learning process, the model acquires knowledge beyond what is typically learned by directly training a classification or retrieval model, and enhances the performance of these tasks. Utilizing generative learning as the mechanism, our method produces heatmaps in a self-supervised manner, eliminating the need for human-annotated heatmaps. Our diffusion-based heatmap generation is inspired by \cite{wolleb2022diffusion}, and we extensively improved the method to handle the co-occurrence of multiple diseases and to generate cleaner and more accurate outputs.

In addition, different from existing work which trains retrieval and classification models separately, we are motivated by the synergy of learning two tasks together. Particularly, the text labels for retrieval provide richer textual context whereas the class labels have a well-defined discrete format. Note that the text labels often omit negative medical findings (i.e., diseases that are not found won't be mentioned) but the class labels always state these explicitly. Therefore, we design an Image-Text-Class Hybrid Contrastive learning criterion which contrasts both image-text and image-class pairs during training. This method not only improves the performance on each task as it leverages both text and class labels, but also results in a single model that supports both classification and retrieval tasks, making real-world deployment at ease.

To validate the proposed two methods, we combine both DAug and the Image-Text-Class Hybrid Contrastive loss and evaluate on the largest Chest X-ray dataset, MIMIC-CXR \cite{johnson2019mimiccxr}. Our model outperforms existing state of the arts on both retrieval and classification tasks.

To sum up, the contribution of this paper is three-fold:
\begin{itemize}
  \item We propose DAug, a portable feature augmentation method which improves medical image understanding performance by adding abnormality heatmaps as additional image channels.
  \item We introduce Image-Text-Class Hybrid Contrastive learning which leverages both image-text and image-class labels to improve performance on both retrieval and classification tasks.
  \item We deliver a single model which is capable of both retrieval and classification tasks with state-of-the-art performance. The proposed methods can be applied to standard pretrained models (such as CLIP ViT \cite{radford2021learning}), making real-world deployment at ease.
\end{itemize}

\section{Related Work}

\subsection{Generation for Perception}
Perception and generation models have long been regarded as two distinct paradigms in machine learning. In computer vision, perception tasks such as classification, retrieval, and segmentation require training data of $<$image, ground truth$>$ pairs labeled in a defined format, which often leads to a limitation in the amount of training data. Generative models, on the other hand, reconstruct the original image during training without requiring annotations, enabling the use of larger amounts of image-only training data. In order to generate new samples, image generation models learn the distribution of images $q(\mathbf{x}_0)$, and this knowledge could aid in the understanding of images as well. This insight has led us to explore the use of a diffusion-based generation model in aiding medical image understanding.

Existing works in image understanding have been benefiting from generative models. One stream of work explores using generative models to augment or synthesize training data \cite{park2018data,shin2018medical,frid2018synthetic}. Another stream of work uses generative modeling as a pretraining task. For example, Masked AutoEncoders (MAE) \cite{he2022masked} reconstructs the original image from a partial input during pretraining, and finetunes on downstream classification tasks. In terms of medical images, \cite{wolleb2022diffusion} proposed medical anomaly detection with classifier-guided diffusion. When applied on Chest X-rays, their work produces reasonable anomaly heatmaps for easy cases like pleural effusion, but cannot handle the co-occurrence of multiple diseases reliably. Our work is based on \cite{wolleb2022diffusion}, and we make significant algorithmic improvements to turn the output into a useful feature that aids downstream tasks. Our improvements are detailed in \autoref{sec:daug}.

\subsection{Image Retrieval and Classification}

In the era of transformer models, pretraining on web-scale data boosts the performance on downstream tasks like retrieval and classification. CLIP \cite{radford2021learning} proposes image-text contrastive learning, which aligns the feature spaces of an image encoder and a text encoder. Retrieval methods base on the cosine similarity of CLIP features have become dominant since then. Due to the significant domain gap between medical images and the common scenic images, CLIP shows limited zero-shot performance in the medical domain. A stream of work \cite{wang2022medclip,eslami2021does} aims to fine-tune CLIP on medical image-text data.

In this study, we address the challenges of medical image retrieval and classification from two perspectives, including DAug from the feature augmentation perspective and the Image-Text-Class Hybrid Contrastive learning from the loss function perspective. Related to feature augmentation, previous studies \cite{wang2022cross, chen2022cross} show that maintaining feature banks for prototypes built on the whole dataset is often effective for tasks with limited training data, and this applies to medical images. Recently, X-TRA \cite{van2023x} proposes improving radiology multi-modal retrieval and classification by a retrieval-based feature augmentation. They use a CLIP model to select the top-K similar samples from the dataset and construct an augmented feature for each input sample. Differently, DAug, our method, leverages generative models and outperforms existing methods.

Related to our Image-Text-Class Hybrid Contrastive loss, UniCL \cite{yang2022unified} also combines image-text and image-class datasets for contrastive pretraining. However, each training sample is assumed to have either a text label or a class label, but not both. This assumption is different from the medical domain, where datasets such as MIMIC-CXR and CheXpert \cite{johnson2019mimiccxr, johnson2019mimicjpg, smit2020chexbert} often contain both text (medical report) and class (multiple disease classes) labels at the same time. Our Image-Text-Class Hybrid Contrastive loss function utilizes this observation by combining both image-text contrastive loss and image-class contrastive loss into a single loss term per sample. Intuitively, the method treats image classification as a special image-to-class retrieval task. It not only enables the synergy of learning both retrieval and classification tasks together, but also delivers a single model that can perform both two tasks. Therefore, both model performance and deployment ease are improved.

\section{DAug: Diffusion-based Feature Augmentation}
\label{sec:daug}

We propose a unified model for multi-modal retrieval and image classification in the radiology domain. Two proposed methods improve the input feature and the training criterion, respectively, and are both portable to a suite of pretrained transformer architectures. This section walks through Diffusion-based Feature Augmentation (DAug) in detail. We first introduce the training of the classifier which guides diffusion. Then, we describe the use of a diffusion-based generative model to create heatmaps and the methods of using these additional features. 

\subsection{Multi-label Image Classifier}
\label{sec:daug_cls}

The disease classifier is trained separately with the diffusion model to provide guidance on generating an image targeting a disease class. The inputs include a noisy image $\mathbf{x}_t$ and the time step $t$. We use sigmoid activation and binary cross-entropy loss to handle multiple labels per image.

For experiments on both retrieval and classification tasks, we use MIMIC-CXR \cite{johnson2019mimiccxr}, the largest Chest X-ray image-text dataset. As class labels are unavailable on this dataset, we follow existing work to generate them with CheXbert \cite{smit2020chexbert}, a text classification model that converts text into a pre-defined set of 14 disease classes.

\subsubsection{Disease Super-classes}


To support our goal of generating heatmaps to enhance the feature for downstream tasks, we extend previous study \cite{wolleb2022diffusion} from a single disease class that is relatively easy (pleural effusion) to multiple disease classes that develop in different regions. We find that simply training a classifier on the 14 CheXbert classes results in unsatisfying result when using the classifier to guide a diffusion model. This is because some of the 14 defined disease classes may share the same or similar visual features. For example, the class ``Consolidation'' includes another class ``Pneumonia'', and ``Pneumonia'' has similar visual features with ``Edema'', although they are triggered by different causes. In sum, the definition of 14 CheXbert classes includes non-visual clues. Therefore, it hurts the performance of image generation when the classifier is forced to distinguish between them. In addressing these challenges, we select representative classes and group them into seven super-classes, and trained the classifier on the super-classes. For example, ``Pleural Effusion'' and ``Pleural Other'' are grouped together to generate a heatmap on abnormalities near the pleural areas. The rationale for using super-classes is two-fold: it eliminates ambiguity in distinguishing between a parent class and its child classes, and it acknowledges that some diseases share similar visual features but differ in their causes. This approach enables the classifier to focus on distinguishing visual features rather than the underlying causes. 

The grouping of child classes should be decided based on the dataset and the tasks. Too many or too few super-classes result in under-fitting and over-fitting to the grouped classes, respectively. We find seven super-classes is a balanced choice for the scenarios in this paper. Please find the list of super-classes and the classifier performance in the supplementary material.

\subsection{Classifier-Guided Diffusion Model}

\begin{figure*}[h!]
  \centering
  \includegraphics[width=.8\linewidth]{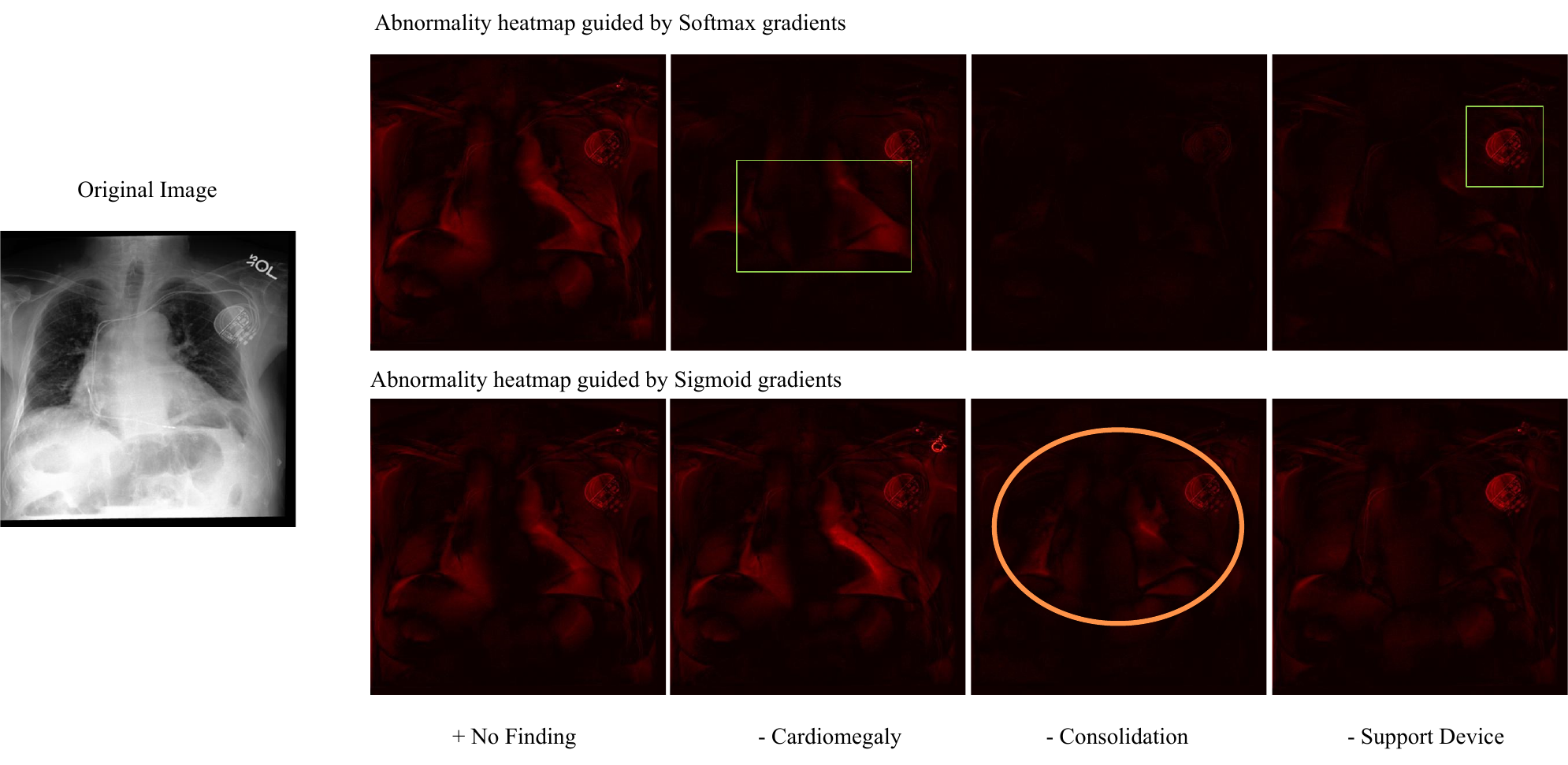}
  \includegraphics[width=.8\linewidth]{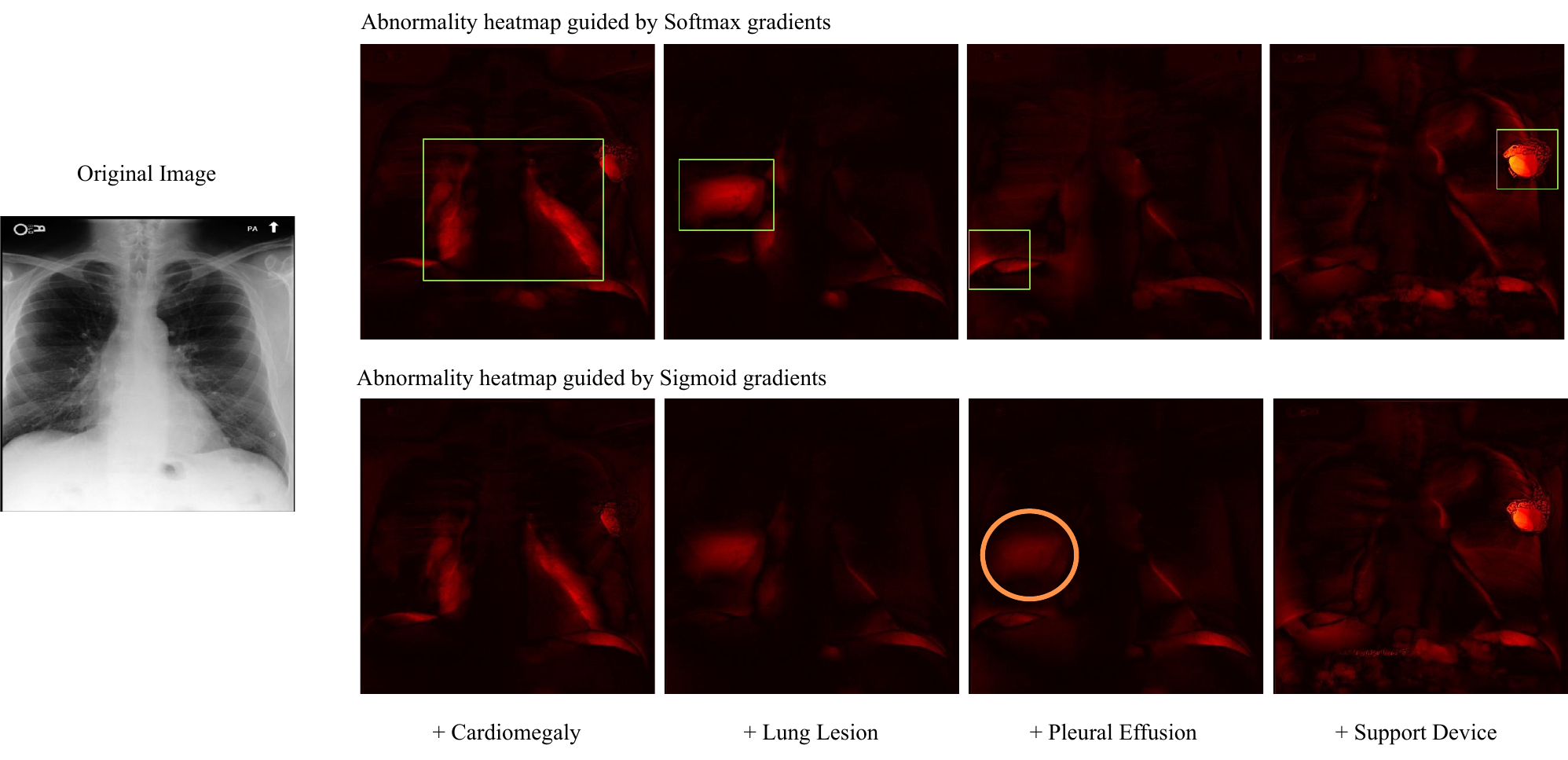}
  \caption{Example output of abnormality heatmaps generated by DAug. We use two chest X-rays as examples, each one covers four types of abnormalities (cardiomegaly, consolidation, etc. in each column). The \textbf{plus }($+$) and \textbf{minus} ($-$) signs indicate the direction of the classifier gradient, meaning amplifying the disease and reducing the disease, respectively. \textbf{``+ No Findings''} reduces the probability of all potential diseases. For each input, the \textbf{first row} shows the output heatmap guided by gradients of \textbf{softmax} probabilities, and the \textbf{second row} shows the results guided by gradients of the \textbf{sigmoid} probabilities. The \textbf{green bounding boxes} shows that our method correctly highlights the region of the disease, which can help the mode to establish better image-text correspondence. Also, using softmax gradient is better than sigmoid gradient as guidance, as softmax successfully removes false positives (see orange circles). \textbf{Orange circle} in the second row highlights false positive areas of consolidation, and the orange circle in the last row highlights a wrong activation of lung lesion when it is supposed to detect pleural effusion. The corresponding softmax version (green box) makes correct detections.}
  \label{fig:heatmap}
\end{figure*}

We train a Denoising Diffusion Probabilistic Model (DDPM) model \cite{ho2020denoising} on Chest X-ray images. Given an input image $I$, we generate a series of noisy images \{$x_0, x_1, \dots, x_T$\} by gradually adding Gaussian noise for $T$ steps. $x_0$ is the original image with no noise and $x_T \sim \mathcal{N}(\mathbf{0}, \mathbf{I})$. A U-Net model is trained to reverse the process by estimating the noise added in any timesteps. During evaluation, the model can recover the original image $x_0$ by removing the estimated noise for $T$ steps from Gaussian noise $X_T$. We set $T=1,000$.

To translate a radiology image into its healthy or diseased variant, the generated image must retain the same anatomic structure as the input image. For this purpose, following \cite{wolleb2022diffusion}, we add noise by only 500 steps to the original input, resulting in $x_{500}$ where the anatomic structures are still visible. The diffusion model is used to recover the original input from $x_{500}$. To produce a diseased or healthy version of the input, the denoising process is guided by the image classifier in \autoref{sec:daug_cls} that assigns an X-ray to defined disease classes. The gradients for a selected class are used to condition the denoising process, making the output image maximize or minimize the probability of the class in the classifier.

For Chest X-ray images, each image may contain multiple diseases, making the binary classifier from the single-class baseline \cite{wolleb2022diffusion} unsuitable. From our evaluation, particularly challenging cases involve diseases that often co-occur (e.g., pleural effusion and lung opacity). In such instances, the generated heatmap highlights areas for both diseases without accurately distinguishing them. This problem arises because the image classifier cannot differentiate diseases that frequently co-exist, where the presence of one disease serves as a strong bias for the presence of the other. To address these challenges, we propose the improvements below.

\subsubsection{Sigmoid for Training and Softmax for Testing}

When using the classifier as guidance, we employ sigmoid activation for training the classifier but softmax for guiding the diffusion model during testing. As multiple diseases can co-exist on a medical image, we train the classifier with sigmoid activation to use multiple disease class labels. During inference, however, we observed that the generated outputs using sigmoid gradients as de-noising guidance tend to highlight false positive regions. This issue can be intuitively explained with an example. Suppose that we are generating a diseased version of the input image where the disease class is cardiomegaly. We want the generated output to contain only cardiomegaly so that the difference heatmap will accurately highlight this particular disease. During the denoising process, we update the noisy image toward a direction that increases the sigmoid probability $z_i$ for cardiomegaly. As the sigmoid equation in \autoref{eq:sigmoid} considers only one class, the generated output is not guaranteed to be free of other diseases (i.e., minimized $z_j, j \neq i$). In fact, as some diseases tend to co-exist (e.g., cardiomegaly and pleural effusion), the generated image tends to have both diseases, which introduces false positives. Using softmax as the activation during inference solves this issue because mathematically \autoref{eq:softmax} enforces maximized $z_i$ while minimizing probability of other classes ($z_j, j \neq i$). Intuitively, the task transforms from ``generating an image where the chance of cardiomegaly is maximized'' to ``generating an image where the chance of cardiomegaly, compared to other diseases, is maximized''. We validate the effectiveness with qualitative comparisions, as shown in \autoref{fig:teaser} (bottom row) and \autoref{fig:heatmap}.

\begin{equation}
  \text{sigmoid}(z_i) = \frac{1}{1+e^{-z_i}} \label{eq:sigmoid}
\end{equation}

\begin{equation}
  \text{softmax}(z_i) = \frac{e^{z_i}}{\sum_{j=1}^{K}e^{z_j}} = \frac{1}{\sum_{j=1}^{K}e^{z_j-z_i}} \label{eq:softmax}
\end{equation}




\begin{figure}[t!]
  \centering
  \includegraphics[width=.7\linewidth]{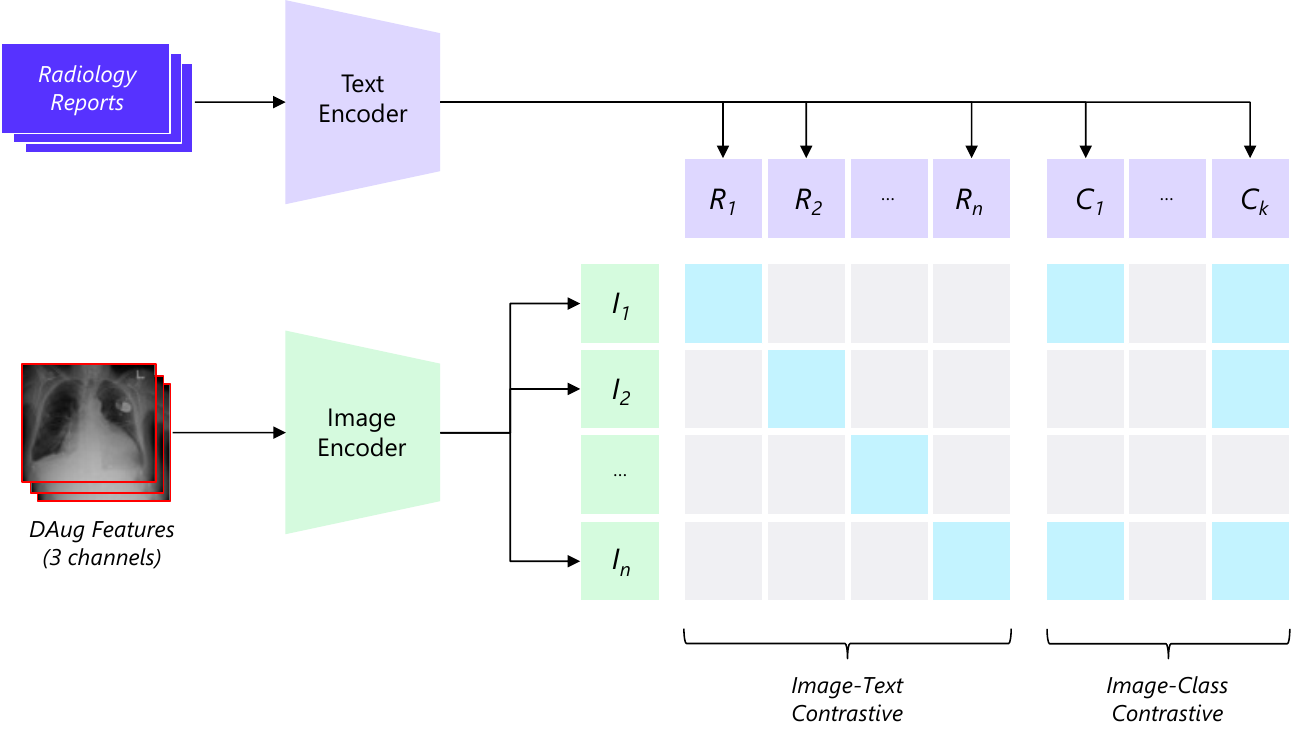}
  \caption{Model architecture and Image-Text-Class Hybrid Contrastive Loss. The inputs are pairs of radiology reports and DAug features (3-channel images including both medical image and abnormality heatmap channels). The image and text encoders are pretrained by CLIP. $R$ and $C$ are text embeddings for the reports and class prompts, respectively. The hybrid contrastive loss includes both the image-text CLIP loss and image-class contrastive loss. Blue cells are positive pairs, in which the image-class matchings are derived from ground-truth class labels.}
  \label{fig:pipeline}
\end{figure}

\begin{table*}[t]
  \centering
  \resizebox{1\textwidth}{!}{
    \begin{tabular}{llllcccccccccccccccc}
      \toprule
      &&&& \rot{No Finding} & \rot{Enl. Cardiomed.} & \rot{Cardiomegaly} & \rot{Lung Opacity} & \rot{Lung Lesion} & \rot{Edema} & \rot{Consolidation} & \rot{Pneumonia} & \rot{Atelectasis} & \rot{Pneumothorax} & \rot{Pleural Effusion} & \rot{Pleural other} & \rot{Fracture} & \rot{Support Devices} & Avg  & \textbf{wAvg}   \\\cmidrule(){1-4}\cmidrule(rl){5-18}\cmidrule(rl){19-19}\cmidrule(l){20-20}
      
      Zhang et al.~\cite{zhang2022category}&\multirow{5}{*}{$r$}&\multirow{5}{*}{$\rightarrow$}&\multirow{5}{*}{$x$}&-&-&-&-&-&-&-&-&-&-&-&-&-&-&.485&-\\
      X-TRA \it {CLIP ($\mathcal{L}_{CLIP}$)} &  &  &  & .62 & .52 & .93 & .88 & .50 & .60 & .29 & .44 & .75 & .54 & .85 & .50 & .36 & .71 & .723 & .606 \\
      X-TRA \it {CNN+BERT}    &  &  &  & \textbf{.77} & .54 & .73 & .91 & .52 & \textbf{.83} & .39 & \textbf{.87} & .77 & .63 & .74 & .23 & .61 & .73 & .735 & .662 \\
      X-TRA \it {PubmedCLIP}  &  &  &  & .75 & .65 & .92 & \textbf{.99} & .23 & .79 & .21 & .51 & .59 & \textbf{.72} & .81 & .56 & .43 & .67 & .720 & .645  \\
      X-TRA \it {CLIP}        &  &  &  & .63 & .62 & \textbf{.96} & .94 & .62 & .69 & .47 & .61 & \textbf{.85} & .69 & \textbf{.91} & .57 & .46 & \textbf{.82} & \textbf{.779} & .703 \\
      \textbf{DAug \it{CLIP} (ours)}        &  &  &  & \textbf{.77} & \textbf{.86} & .83 & .89 & \textbf{.64} & .79 & \textbf{.80} &.71 &	.81 &	.64 &	.79 &	\textbf{.66} & \textbf{.76} &	.79 & .767 & \textbf{.799} \\
      \cmidrule(){1-4}\cmidrule(rl){5-18}\cmidrule(rl){19-19}\cmidrule(l){20-20}
      Yu et al.~\cite{yu2021multimodal}&\multirow{5}{*}{$x$}&\multirow{5}{*}{$\rightarrow$}&\multirow{5}{*}{$x$}&-&.65&.75&.72&.43&.80&.73&.60&.76&.76&.85&.43&.16&.86&.680&-\\
      X-TRA\cite{van2023x} \it{CLIP ($\mathcal{L}_{CLIP}$)}&&&&  .71 & .52 & .74 & .78 & .39 & .79 & .39 & .40 & .76 & .42 & .67 & .44 & .43 & .64 & .761 & .578 \\
      X-TRA \it{CNN+BERT}    &  &  &  & .87 & .63 & .88 & \textbf{.90}  & .49 & .90  & .57 & .60  & .85 & \textbf{.85} & .83 & .29 & .47 & .82 & .769 & .678 \\
      X-TRA \it{PubmedCLIP}  &  &  &  & \textbf{.90}  & .63 & .82 & .83 & .39 & .86 & .45 & .63 & .87 & .53 & \textbf{.90}  & .48 & .51 & .79 & .795 & .685 \\
      X-TRA \it{PubmedCLIP}  &  &  &  & \textbf{.90}  & .63 & .82 & .83 & .39 & .86 & .45 & .63 & .87 & .53 & \textbf{.90}  & .48 & .51 & .79 & .795 & .685 \\
      X-TRA \it{CLIP}        &  &  &  & .84 & .62 & \textbf{.89} & .89 & \textbf{.56} & \textbf{.91} & .55 & .59 & \textbf{.89} & .60  & .86 & .49 & .57 & \textbf{.84} & \textbf{.840} & .713  \\
      \textbf{DAug \it{CLIP} (ours)}        &  &  &  & .72 & \textbf{.86} & .83 & .88 & .53 & .78 & \textbf{.79} & \textbf{.65} & .81 & .47 & .77 & \textbf{.59} & \textbf{.69} & .74 & .721 & \textbf{.771} \\
      \cmidrule(){1-4}\cmidrule(lr){5-18}\cmidrule(lr){19-19}\cmidrule(l){20-20}

      
    \end{tabular}
  }
  \caption{\textbf{Retrieval performance} for both report-to-image (r-x) and image-to-image(x-x) scenarios, measured with retrieval mAP@K, where K=5. wAvg and Avg are the weighted average of classes and the average of classes, respectively. \textbf{wAvg} is the primary metric as long-tailed challenges is not a focus of this paper. Our model, DAug-CLIP surpasses existing methods by a clear margin in the prevalent report-to-image scenario. In terms of image-to-image retrieval, DAug-CLIP achieves state-of-the-art performance on the most common classes and in terms of wAvg. This could be attributed to the reliability of the DAug feature in these classes, as the classifier guiding the diffusion model is affected by class imbalance.}
  \label{tab:retrieval_main}
  \end{table*}        

    \begin{table*}[t!]
      \centering    
      \resizebox{1\textwidth}{!}{
      \begin{tabular}{lccccccccccccccccc}
        \toprule
                            &  Aug                & \rot{No Finding} & \rot{Enl. Cardiomed.} & \rot{Cardiomegaly} & \rot{Lung Opacity} & \rot{Lung Lesion} & \rot{Edema} & \rot{Consolidation} & \rot{Pneumonia} & \rot{Atelectasis} & \rot{Pneumothorax} & \rot{Pleural Effusion} & \rot{Pleural other} & \rot{Fracture} & \rot{Support Devices}  & Avg  & \textbf{wAvg}  \\\cmidrule(r){1-2}\cmidrule(rl){3-16}\cmidrule(rl){17-17}\cmidrule(rl){18-18}
      
      CNN+BERT& - &.81 & .63 & .73 & .67 & .62 & .83  & .69 & .59 & .68 & .75  & .83 & .70  & .58 & .84 & .79 & .71 \\
      &X-TRA\cite{van2023x}&.81 & .74 & .75 & .69 & .63 & .81  & .72 & .63 & .75 & .75  & .83 & .69  & .63 & .85& .82 & .73  \\\cmidrule(r){1-2}\cmidrule(rl){3-16}\cmidrule(rl){17-17}\cmidrule(rl){18-18}
      
      PubmedCLIP&-&.78 & .65 & .72 & .66 & .61 & .82  & .70 & .61 & .73 & .76  & .81 & .62  & .54 & .84 & .78 & .70 \\
      &X-TRA\cite{van2023x}&.84 & .76 & .78 & .69 & .64 & .83  & .73 & .64 & .76 & .75  & .82 & .75  & .67 & .85 & .83 & .75 \\\cmidrule(r){1-2}\cmidrule(rl){3-16}\cmidrule(rl){17-17}\cmidrule(rl){18-18}
      
      &-&.77 & .65 & .71 & .67 & .62 & .85  & .73 & .61 & .72 & .75  & .80 & .59  & .51 & .83 & .80 & .70 \\
      CLIP&X-TRA\cite{van2023x}&.82 & .78 & .74 & .70 & .71 & .82  & .75 & .63 & .79 & .78  & .86 & .74  & .72 & .91 & .85 & .77 \\
      &\textbf{DAug, ours}  & \textbf{.93} & \textbf{.90} & \textbf{.90} & \textbf{.91} & \textbf{.85} & \textbf{.89} & \textbf{.89} & \textbf{.85} & \textbf{.89} & \textbf{.90} & \textbf{.90} & \textbf{.87} & \textbf{.84} & \textbf{.89} & \textbf{.89} & \textbf{.89} \\
      \bottomrule
      \end{tabular}%
      }
      \caption{\textbf{Image classification} performance on MIMIC-CXR dataset. Results are in AUC-ROC. wAvg is the weighted average by number of samples per class. Avg is the average. Augmentation method X-TRA improves baseline performance, with CLIP achieving the best result. DAug-CLIP, our method, surpasses the best setting of X-TRA by a clear margin in all classes.}
      \label{tab:main_cls}
      \end{table*}

  \subsubsection{Channel-wise Feature Augmentation}
  \label{sec:fea_aug}
  The generated heatmaps can be integrated into the downstream perception model in various ways. A simple method involves processing them through a vision backbone and then concatenating or adding the heatmap features to the image features for downstream tasks. This straightforward approach necessitates modifications to the model architecture. We propose an alternative, channel-wise feature augmentation method, which incorporates the heatmaps as additional image channels alongside the medical image.

  This channel-wise feature augmentation offers two main advantages. First, it leverages the capabilities of powerful pretrained vision models designed for RGB three-channel input. As radiology images are typically monochrome, the channel-wise augmentation makes better use of the computational budget for three channels. Second, channel-wise augmentation does not require any changes to the model architecture, making it easier to utilize a wide range of pretrained transformers. This approach is complementary to other methods, enhancing performance on downstream tasks without the need for architectural modifications.

%
%

\begin{table*}[t]
  \centering
  \resizebox{1\textwidth}{!}{
    \begin{tabular}{lcccccccccccccccccc}
      \toprule
      Method &  Feature Aug & Criterion & \rot{No Finding} & \rot{Enl. Cardiomed.} & \rot{Cardiomegaly} & \rot{Lung Opacity} & \rot{Lung Lesion} & \rot{Edema} & \rot{Consolidation} & \rot{Pneumonia} & \rot{Atelectasis} & \rot{Pneumothorax} & \rot{Pleural Effusion} & \rot{Pleural other} & \rot{Fracture} & \rot{Support Devices}  & Avg & \textbf{wAvg}   \\\cmidrule(r){1-1}\cmidrule(r){2-2}\cmidrule(r){3-3}\cmidrule(rl){4-17}\cmidrule(rl){18-18}\cmidrule(rl){19-19}

      \rowcolor{lightgray}
      \textit{Image Retrieval: $r \rightarrow x$ \vspace{1mm}} \\
      X-TRA (Baseline) & X-TRA & CLIP & .62 & .52 & .93 & .88 & .50 & .60 & .29 & .44 & .75 & .54 & .85 & .50 & .36 & .71 & .723 & .606 \\
      DAug (Ablation) & DAug & CLIP & \textbf{.77} & .85 & .80 & .86 & .60 & .76 & .78 & .67 & .79 & .61 & .76 & .65 & .75 & \textbf{.79} & .745 & .778 \\
      DAug (Full) & DAug & Hybrid Contrastive & \textbf{.77} & \textbf{.86} & \textbf{.83} & \textbf{.89} & \textbf{.64} & \textbf{.79} & \textbf{.80} & \textbf{.71} &	\textbf{.81} &	\textbf{.64} &	\textbf{.79} &	\textbf{.66} & \textbf{.76} &	\textbf{.79} & \textbf{.767} & \textbf{.799} \\
      \cmidrule(r){1-1}\cmidrule(r){2-2}\cmidrule(r){3-3}\cmidrule(rl){4-17}\cmidrule(rl){18-18}\cmidrule(rl){19-19}

      \rowcolor{lightgray}
      \textit{Image Retrieval: $x \rightarrow x$ \vspace{1mm}} \\
      X-TRA (Baseline) & X-TRA & CLIP & .71 & .52 & .74 & .78 & .39 & .79 & .39 & .40 & .76 & .42 & .67 & .44 & .43 & .64 & .761 & .578 \\
      DAug (Ablation) & DAug & CLIP & .71 & \textbf{.86} & .81 & \textbf{.88} & .49 & .76 & .78 & .62 & \textbf{.81} & .44 & .76 & .58 & \textbf{.69} & \textbf{.74} & .711 & .765 \\
      DAug (Full) & DAug & Hybrid Contrastive & \textbf{.72} & \textbf{.86} & \textbf{.83} & \textbf{.88} & \textbf{.53} & \textbf{.78} & \textbf{.79} & \textbf{.65} & \textbf{.81} & \textbf{.47} & \textbf{.77} & \textbf{.59} & \textbf{.69} & \textbf{.74} & \textbf{.721} & \textbf{.771} \\

      \cmidrule(r){1-1}\cmidrule(r){2-2}\cmidrule(r){3-3}\cmidrule(rl){4-17}\cmidrule(rl){18-18}\cmidrule(rl){19-19}
      \rowcolor{lightgray}
      \textit{Image Classification \vspace{1mm}} \\
      CLIP (Baseline) &- & CLIP &.77 & .65 & .71 & .67 & .62 & .85  & .73 & .61 & .72 & .75  & .80 & .59  & .51 & .83 & .80 & .70 \\
      X-TRA (Baseline) &X-TRA & CLIP & .82 & .78 & .74 & .70 & .71 & .82  & .75 & .63 & .79 & .78  & .86 & .74  & .72 & .91 & .85 & .77 \\
      DAug (Ablation) & DAug & CLIP & .77 & .86 & .83 & .89 & .64 & .79 & .80 &.71 &	.81 &	.64 &	.79 &	.66 & .76 &	.79 & .77 & .80 \\
      DAug (Full) & DAug & Hybrid Contrastive & \textbf{.93} & \textbf{.90} & \textbf{.90} & \textbf{.91} & \textbf{.85} & \textbf{.89} & \textbf{.89} & \textbf{.85} & \textbf{.89} & \textbf{.90} & \textbf{.90} & \textbf{.87} & \textbf{.84} & \textbf{.89} & \textbf{.89} & \textbf{.89} \\
    \bottomrule
    \end{tabular}%
  }\caption{\textbf{Ablation studies} on image retrieval and classification tasks. Compared to baselines, our feature augmentation method DAug gains performance over no augmentation (CLIP Baseline) and the X-TRA augmentation (X-TRA Baseline). Besides, using the proposed Image-Text-Class Hybrid Contrastive loss outperforms the baselines using the original CLIP loss. Baseline results from \cite{van2023x}.}
  \label{tab:ablation}
\end{table*}

\section{Image-Text-Class Hybrid Contrastive Learning}
  \label{sec:model}

  Image-to-text retrieval and image classification are deeply interconnected tasks. Essentially, image classification can be seen as a retrieval problem focusing on a smaller, reused set of targets. When the classification head is a linear layer without bias, the class logits become unscaled cosine similarities between the image feature and the weights in the linear layer. In image-to-text retrieval, the weights in the linear layer are replaced by text embeddings dynamically generated for each target text. Motivated by the potential benefits of jointly training both retrieval and classification tasks, we integrate image-text and image-class labels into our training loss. Unlike existing methods \cite{khosla2020supervised,yang2022unified} that expand contrastive learning to class labels, our approach uniquely addresses scenarios where a single sample is associated with both text and class labels, aiming to train a unified model for retrieval and classification. To distinguish from existing work, we name our method image-text-class hybrid contrastive learning.

  As illustrated in \autoref{fig:pipeline}, we first transform each class into a fixed set of texts by converting each class into prompts. For instance, the class "Cardiomegaly" is rephrased as "A photo of a Chest X-ray image with cardiomegaly". During training, as depicted in \autoref{fig:pipeline}, class prompts and radiology reports are transformed into text embeddings $C$ and $R$, respectively. We then calculate two sets of losses and take the weighted average:

  \begin{equation}
    \mathcal{L} = w * \mathcal{L}_{CLIP} + (1-w) * \mathcal{L}_{i2c},
    \label{eq:loss}
  \end{equation}

  where $\mathcal{L}_{CLIP}$ is the CLIP loss including both image-to-text and text-to-image cross-entropy loss, and $\mathcal{L}_{i2c}$ is the image-to-class binary cross-entropy loss. Specifically,

  \begin{equation}
    \mathcal{L}_{i2c} = -\frac{1}{N} \sum_{i=1}^{N} \left[ y_i \log(\text{sim}(i, j)) + (1 - y_i) \log(1 - \text{sim}(i, j)) \right],
    \label{eq:bce}
  \end{equation}

  where $\text{sim}(i, j)$ is the cosine similarity between the image embedding $I_i$ and text embedding for class prompts $C_i$. During training, embedding $C_i$ is regenerated with each update to the text encoder. Several works discussed the connection between contrastive learning and cross-entropy loss \cite{khosla2020supervised,yang2022unified}. In our scenario, $\mathcal{L}_{i2c}$ is essentially contrasting image $I$ with the class prompts, where multiple positive pairs could exist determined by the ground truth class labels. The inherent cross-entropy nature of both loss terms facilitates training both tasks simultaneously without causing the embedding space to diverge for each task.

  \subsection{Multi-modal Retrieval and Classification}
  \label{sec:tasks}

  After the model is trained, we leverage the model in different ways for the classification and retrieval tasks. For classification, we use only the image encoder and connect it with a linear classifier. The linear classifier consists of a single layer where the weight vectors per class are populated with the text embeddings for class prompts generated by the final text encoder. The bias is set to zeros. Essentially, the output logit per class is equivalent to the unscaled cosine similarity between the image and classes $\text{logit}_i = sigmoid(I \cdot C_i)$.

  For retrieval tasks, both the image and text encoders are used to convert each sample into embeddings. We use the cosine similarity of the embeddings to rank the association for retrieval.

\section{Experiments}
\subsection{Implementation details}

To benchmark our ideas, we need an image-text-class dataset on medical images. We select \textbf{MIMIC-CXR}, the largest Chest X-ray (CXR) medical report dataset. It contains 227,835 image-text pairs, where the texts are radiology reports which list the normal and abnormal findings. We follow the official training and testing splits. As class labels are unavailable, we generate pseudo-class labels with the CheXbert labeler \cite{jain2021visualchexbert}, which is a standard practice in the field. It is a text classification model that converts a radiology report into binary labels on 14 disease classes. One of them is ``No Findings'', indicating a healthy case.

For DAug feature augmentation, we construct a three-channel image with the first two channels containing the medical image and the third channel filled with the diffusion-generated heatmap guided by the class ``No Findings''. In our experiments, we evaluate all 14 disease classes and compare the results with the original CLIP as a baseline. Therefore, we select the heatmap for ``No Finding'' which combines all diseases. This requires no change to the model architecture for a fair comparison with the vanilla CLIP. In real-world applications, heatmaps for individual disease groups (e.g., cardiomegaly) can be selected for optimal performance gain according to the scenario.

Following existing work \cite{van2023x}, we resize images to $256 \times 256$ and use a CLIP pretrained ViT-Base/32 model for fair comparison. We fine-tune the model with a cosine learning rate scheduler with a base learning rate of $2e^{-5}$ for 10 epochs. We set the weight $w$ in \autoref{eq:loss} to 0.7 based on the validation performance of two tasks. We use a batch size of 256 over eight V-100 GPUs. With all images resized in advance, training and evaluation take around 2 hours. Abnormality heatmaps are pre-generated.

\subsection{Results}
We compare our method, DAug with existing methods on both multi-modal retrieval and image classification on the MIMIC dataset. \autoref{tab:retrieval_main} demonstrates that our method outperforms existing state-of-the-art approaches in retrieving radiology images with medical reports, a critical clinical scenario where radiologists refer to previous cases to confirm diagnoses. \autoref{tab:main_cls} shows the performance of the same model on image classification, which outperforms existing methods by a clear margin. Specifically, we find that X-TRA benefits more in the tail classes, while DAug improves the overall performance which is reflected in the higher wAvg score.


DAug produces heatmaps without requiring ground truth. Unfortunately, the lack of ground truth also constrains us on measuring the quality of the heatmaps. Considering that our end goal is to improve the performance of downstream tasks, we validate the heatmap quality by measuring the improvement of these tasks when using the heatmaps. Therefore, we conduct ablation studies in \autoref{tab:ablation}. The results show that performance on both tasks surpasses the baseline when only changing the feature augmentation approach to DAug, which proves that the heatmaps are helpful features. When both the DAug feature augmentation and the proposed Hybrid Contrastive criterion are used, the performance on both tasks is improved. This comparison validates that both methods effectively aid in medical image understanding tasks.



\section{Conclusion}
In this paper, we present a single model that achieves state-of-the-art performance in both medical image retrieval and classification tasks. We propose DAug, a diffusion-based channel-wise feature augmentation method that empirically directs the model where to look for clinical diagnoses. We enable multi-class support and reduce the false positive issue found in previous work by using super-classes and softmax gradients for diffusion guidance. In addition, we propose the Image-Text-Class Hybrid Contrastive learning criterion, which leverages both image-text and image-class labels and enables a unified model for two tasks with easy deployment. Our method requires no modifications to the model architecture, making it portable to a wide range of pretrained models.




\bibliographystyle{ieeetr}
\bibliography{daug}

\newpage


\appendix

\section{Details on Image Generation and Image-to-Image Translation}
\label{sec:diffusion}
Common image generation methods include Generative Adversarial Networks (GANs) \cite{goodfellow2020generative}, Variational Autoencoders (VAEs) \cite{kingma2013auto} and Diffusion Models \cite{song2019generative}. Diffusion models have become mainstream due to the ease of training and the superior image quality. During training, the diffusion model learns to remove noise from a noisy input and therefore can gradually turn a Gaussian noise into an image. Such denoising steps can be guided by an image classifier trained separately, whose gradients are used to determine the direction of the denoising process, encouraging the output to maximize the probability of a certain class based on the classifier. The result will be an image of the chosen class.

To train a diffusion model, we first conduct a forward diffusion process which gradually adds Gaussian noise to the original image. The forward process has $T$ steps (usually, $T=1000$), producing a sequence of noisy samples $\mathbf{x}_1, \dots, \mathbf{x}_T$. $\mathbf{x}_0=\mathbf{I}$ is the original image and $\mathbf{x}_T \sim \mathcal{N}(\mathbf{0}, \mathbf{I})$ becomes a Gaussian noise. Then, a U-Net model is trained to predict the noise added per step, in order to reverse the steps by removing the added noise from the noisy image. The reverse process gradually turns a Gaussian noise into an image. Diffusion models can be guided by an image classifier to generate an image of a particular class. The image classifier is trained to produce class probabilities given a noisy image $f_\phi(y | \mathbf{x}_t, t)$. The gradients of the classifier are used to alter the denoising at each step so that the output maximizes the probability of the target class.

A classifier-guided diffusion model can be used for image-to-image translation. Specifically, we obtain a half-noised image at time step $x=500$, and then conduct denoising guided by a classifier. As $\mathbf{x}_{500}$ maintains key distinguishable features of the original image, the output still maintains the identity of the original input, but changes it in a way that will be classified to the target class. Related to medical images, this is about converting a healthy CXR to a diseased one, and vice versa.

\section{Disease Super-classes}

In our classifier-guided diffusion model, the classifier is trained on disease super-classes, with each super-class consisting of one or multiple related diseases.
The super-classes were defined with radiologists to align with medical knowledge. For the additional rationale behind this definition, CheXpert \cite{smit2020chexbert} provides a hierarchical structure of the 14 disease classes, which aligns with our super-class definition. Take super-class \#4 as an example, they are grouped together because they are shown as increased density in the X-ray, although for different reasons. The goal is to let the classifier focus only on the appearance instead of attempting to distinguish the cause. For example, in \cite{smit2020chexbert}, Atelectasis is another type of lung opacity abnormality. We categorize it separately because it looks different (an absence of density).

We show the classifier performance on the super-classes in \autoref{tab:cls_perf}. Upon empirical examinations, we found that the quality of the heatmaps is highly correlated with the classifier performance of the selected class. This observation supports the decision to group sub-classes together to improve the classifier's robustness. Please note that the performance in the table is expected to be low, as input to the classifier is noisy images instead of the original image (see \autoref{sec:diffusion} for details about classifier-guided diffusion).

\begin{table}
  \centering
  \setlength{\tabcolsep}{8pt} 
  \begin{tabularx}{0.9\textwidth}{clc}
      \toprule
      Super Class & Disease classes & AP\\
      \midrule
      1 & No Finding & $0.631$\\
      2 & Enlarged Cardiomediastinum, Cardiomegaly & $0.885$\\
      3 & Lung Lesion & $0.407$\\
      4 & Consolidation, Edema, Pneumonia & $0.857$\\
      5 & Atelectasis & $0.778$\\
      6 & Pleural Effusion, Pleural Other & $0.746$\\
      7 & Support Device & $0.633$\\
      \bottomrule
  \end{tabularx}
  \caption{Multi-label classifier performance in Average Precision (AP). Please note that this is the classifier that takes in the noisy image and is trained to guide the diffusion model. There are totally seven classes, and each one is a super-class consisting of disease classes with similar visual features. For example, super class 4 includes Edema and Pneumonia, which are sub-categories of Consolidation. Training the classifier with merged classes reduces class imbalance and improves performance, and therefore provides better guidance for the diffusion model.}
  \label{tab:cls_perf}
\end{table}

\section{Pseudo-label Quality}
The class labels are not human annotated but are generated with CheXpert, a text classification model that converts a radiology report into disease classes. According to human evaluation in \cite{smit2020chexbert}, the label quality is claimed to have a 96.9\% F1 score.

\newpage 

\section{Ethical Considerations and Limitations}
Our use of the MIMIC-CXR dataset was approved through PhysioNet\footnote{https://physionet.org/content/mimic-cxr/2.0.0/}. All authors who accessed the data have obtained permission.

We identify two limitations of our work. First, to be compatible with the pretrained models, we configure the input image to be 3 channels. The method may achieve even better results if heatmaps of all supported super-classes are used. Second, the use of a diffusion model introduces significant computational overhead. To address this limitation, it worth exploring to use the heatmaps as an additional supervision to improve a student model. We will take these directions as our future work.

\end{document}